\documentclass[letterpaper, 10 pt, conference]{ieeeconf} 

\IEEEoverridecommandlockouts

\usepackage{graphicx} 
\usepackage[utf8]{inputenc}
\usepackage{amsmath,amsfonts,booktabs,cite} 
\usepackage{overpic,pict2e} 
\usepackage{siunitx,textcomp} 
\usepackage{algorithmic,algorithm} 
\usepackage[hidelinks]{hyperref} 
\usepackage[caption=false]{subfig} 

\let\labelindent\relax
\usepackage[inline]{enumitem} 

\usepackage[nolist]{acronym} 
\newacro{bn}[BN]{Batch Normalization}
\newacro{relu}[ReLU]{Rectified Linear Unit}
\newacro{adam}[Adam]{Adaptive Moment Estimation}
\newacro{ai}[AI]{Artificial Intelligence}
\newacro{dl}[DL]{Deep Learning}
\newacro{dnn}[DNN]{Deep Neural Network}
\newacro{bnn}[BNN]{Bayesian Neural Network}
\newacro{mc}[MC]{Monte-Carlo}
\newacro{slam}[SLAM]{Simultaneous Localization and Mapping}
\newacro{win}[W]{windows}
\newacro{teb}[T]{tables}
\newacro{doors}[D]{glass doors}

\usepackage{standalone,tikz,pgfplots} 
\pgfplotsset{compat=1.13}

\newcommand{\figref}[1]{\hyperref[#1]{Figure~\ref*{#1}}}
\newcommand{\tabref}[1]{\hyperref[#1]{Table~\ref*{#1}}}
\newcommand{\secref}[1]{\hyperref[#1]{Section~\ref*{#1}}}
\newcommand{\algoref}[1]{\hyperref[#1]{Algorithm~\ref*{#1}}}
\newcommand{\figsref}[2]{Figures~\ref{#1} and \ref{#2}}

\newlength\myindent
\newcommand\bindent{%
	\begingroup
	\setlength{\itemindent}{\myindent}
	\addtolength{\algorithmicindent}{\myindent}
}
\newcommand\eindent{\endgroup}

\newcommand{\matr}[1]{\mathbf{#1}}
\DeclareMathOperator*{\argmax}{arg\,max}

\newcommand{\de}[1]{\operatorname{d}\!#1}

\def\bestcolor{(best viewed in color)}
\def\truedist{robot-to-obstacle distance}
\def\cob{Care-O-bot 4}
\def\sota{state-of-the-art}
\def\valI{Val2F}
\def\valII{Val3F}
\def\drop{\ac{mc} Dropout}

\def\ie{\textit{i.e.,}}
\def\eg{\textit{e.g.,}}

\title{\LARGE \bf%
Deep Network Uncertainty Maps for Indoor Navigation}

\author{Francesco~Verdoja\textsuperscript{*}, Jens~Lundell\textsuperscript{*} 
and Ville~Kyrki%
\thanks{\textsuperscript{*}These authors contributed equally to this 
paper.}%
\thanks{This work was supported by the Strategic Research Council at Academy of 
Finland, decision 314180.}%
\thanks{F.~Verdoja, J.~Lundell and V.~Kyrki are with School of Electrical 
Engineering, Aalto University, Finland. \url{name.surname@aalto.fi}}}

\begin{document}

\maketitle
\thispagestyle{empty}
\pagestyle{empty}


\begin{abstract}
Most mobile robots for indoor use rely on 2D laser scanners for localization, 
mapping and navigation. These sensors, however, cannot detect transparent 
surfaces or measure the full occupancy of complex objects such as tables. Deep 
Neural Networks have recently been proposed to overcome this limitation by 
learning to estimate object occupancy. These estimates are nevertheless subject 
to uncertainty, making the evaluation of their confidence an important issue 
for these measures to be useful for autonomous navigation and mapping. 
In this work we approach the problem from two sides. First we discuss 
uncertainty estimation in deep models, proposing a solution based on a fully 
convolutional neural network. The proposed architecture is not restricted by 
the assumption that the uncertainty follows a Gaussian model, as in the case of 
many popular solutions for deep model uncertainty estimation, such as 
Monte-Carlo Dropout. We present results showing that uncertainty over obstacle 
distances is actually better modeled with a Laplace distribution. 
Then, we propose a novel approach to build maps based on Deep Neural 
Network uncertainty models. In particular, we present an algorithm to build a 
map that includes information over obstacle distance estimates while taking 
into account the level of uncertainty in each estimate. 
We show how the constructed map can be used to increase global 
navigation safety by planning trajectories which avoid areas of high 
uncertainty, enabling higher autonomy for mobile robots in indoor settings.
\end{abstract}


\section{Introduction}
\label{sec:intro}

\acp{dnn} have recently found increased adoption in robotics, where they are 
being proposed for applications such as grasping~\cite{varley_shape_2017} and 
autonomous navigation~\cite{pfeiffer2017perception}. This is, however, raising 
concerns over \ac{ai} safety~\cite{amodei_concrete_2016}, especially in 
situations where unreliable predictions can cause damage to expensive hardware 
or even human harm, as in the case of recent accidents involving autonomous 
vehicles, where in May 2016 the perception system in a car failed to recognize the presence and distance of an obstacle which unfortunately ended in a fatal crash.
In order to integrate \ac{dl} into applications, 
it is therefore important that such systems are able to produce reliable 
estimates of the uncertainty associated with their predictions. 

\begin{figure}
	\centering
	\subfloat[\label{fig:coffee}Coffee 
	room]{\includegraphics[height=13em]{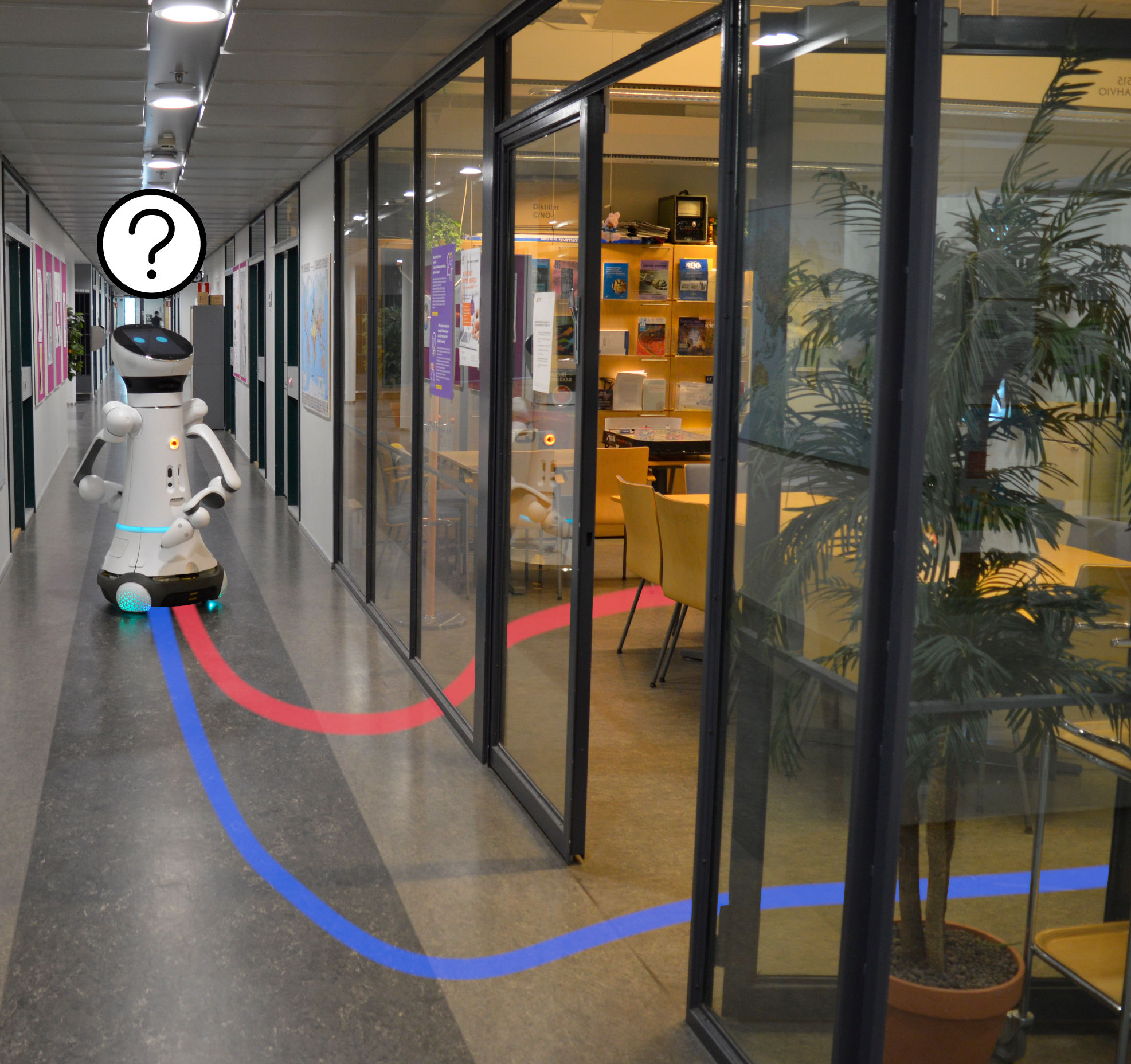}}
	\subfloat[\label{fig:coffee_slam}SLAM 
	map]{\includegraphics[height=13em]{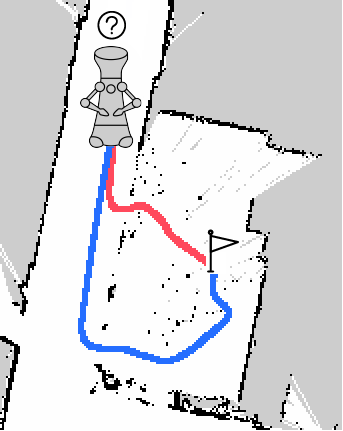}}
	\caption{\label{fig:paths}Two possible navigation trajectories that a robot 
	relying only on 2D lidar could plan to reach a goal. The blue trajectory is 
	longer but safe, while the red one would result in a collision with the 
	glass walls of the room \bestcolor.}
\end{figure}

Autonomous navigation of mobile robots is one of the areas where \ac{ai} safety 
is of utmost importance, both in regards to autonomous cars and to robots 
designed to be employed in industrial settings, since
the robots are often big enough 
to potentially damage the environment or harm the users.
Autonomous cars often employ a wide array of sensors to estimate the 
state of their surroundings while most indoor mobile robots 
rely only on 2D laser scanners for navigation, mapping and 
localization as they provide distance measurements at fast rates and in wide 
angular fields~\cite{baltzakis_fusion_2003}. However, 2D laser scans offer a 
limited amount of information that may be insufficient for tasks like 
object detection and obstacle avoidance. In particular, 2D lidars are not able 
to detect transparent obstacles, \eg{} glass, and are limited to measuring 
occupancy at a single height and therefore cannot infer the true occupancy of 
complex objects such as tables. For example, \figref{fig:paths} shows two
trajectories planned on a map built using 2D lidars only; a robot could not 
determine that the red path would collide with the glass walls of the room 
shown in \figref{fig:coffee} as they are invisible to the laser scanners.

To overcome these limitations, we recently proposed a virtual laser-like 
sensor, using a convolutional network trained to infer, or \emph{hallucinate}, 
the actual distance of objects from raw 2D laser input 
\cite{lundell_hallucinating_2018}. We refer to this distance as the 
\emph{\truedist{}}, \ie{} ``the distance along a certain direction to the 
closest point of an obstacle that the robot could collide with.'' We 
demonstrated in that work the ability of the approach to estimate the occupancy 
of complex objects (\eg{} tables and glass walls), and how it could improve 
safety for local navigation, while its applications to global navigation and 
mapping was left unexplored. 

Using such a sensor for mapping would allow building maps which account for 
actual obstacle occupancy and could yield safer navigation when compared to 
those built by using laser data alone. However, when building occupancy grid 
maps from laser data, the sensor uncertainty is so low to fall inside the grid 
resolution and for that reason is usually neglected. This is not the case for 
\truedist{} estimates: the network uncertainty is too wide to ignore and for 
that reason it has to be accounted for while mapping.

Therefore, in this work we propose to model the uncertainty of the 
convolutional network in~\cite{lundell_hallucinating_2018} and exploit the 
model to build uncertainty maps. \figref{fig:hallucinated_uncertainty_laplace} 
shows an example of such a map, where trajectories can be planned to avoid 
areas of high uncertainty---marked in red---resulting in safer paths.

The most widely adopted approach to predict uncertainty of deep models is known 
as \drop{}~\cite{gal_dropout_2016}. However, it has been shown that the 
application of \drop{} to some domains can be 
problematic~\cite{osband_deep_2016,osband_risk_2016,pearce_bayesian_2018}, and 
we demonstrate how it cannot be used effectively for the task at hand. We then 
propose an alternative \ac{dl}-based method which is able to produce adequate 
prediction of uncertainty. 

The main contributions of this study are:
\begin{enumerate*}[label=(\arabic*)]
	\item a novel \ac{dl}-based approach for uncertainty prediction 
	(\secref{sec:model} and \secref{sec:net}) and its application in the domain 
	of \truedist{} estimation from raw 2D laser data;
	\item an algorithm for building uncertainty maps of indoor environments 
	(\secref{sec:map}) that can be used to integrate uncertainty predictions in 
	global navigation frameworks;
	\item an empirical comparison of the proposed technique with the de-facto 
	standard \drop{} for the task at hand, showing how the proposed approach is 
	a superior choice in this domain (\secref{sec:exp}).
\end{enumerate*}

\section{Related Works}
\label{sec:related}

\subsection{Uncertainty in robotic mobility}
\label{sec:navunc}

In the context of robotic mobility, we recognize three different areas 
where recent research addressed uncertainty: 
\begin{enumerate*}[label=(\arabic*)]
	\item localization, where the goal is to reduce uncertainty regarding the 
	robot position in relation to a known environment;
	\item local navigation, where robot movements should account for sensor 
	uncertainty over obstacle position and occupancy; and
	\item mapping and global navigation, where planned trajectories should avoid
	areas whose content is uncertain, and prefer obstacle-free paths a robot 
	can safely navigate.
\end{enumerate*}
Here, we only review (2) and (3) as they are central for our work, but we 
encourage readers interested in (1) to read, 
\eg{}~\cite{bry_rapidly-exploring_2011,rodriguez-arevalo_importance_2018}.  

\paragraph{Local navigation}
Ensuring safety in the context of local navigation requires to estimate from 
sensory data the position and occupancy of obstacles surrounding the robot. 
To achieve this task robots often rely on raw 2D laser data, despite
laser sensors inherent limitation in providing correct distance 
estimates for many complex objects, including tables, chairs, windows, and open 
shelfs. A recent paper~\cite{axelrod_provably_2017} presented a method for 
planning theoretically safe paths in environments sensed by a 2D lidar. The 
authors proposed to inflate obstacles by a volume that represented both the 
pose uncertainty and the space the obstacle may occupy. However, the 
experimental demonstration was limited to simple objects and nothing was advised for tackling more complex objects. Another option to reduce obstacle 
uncertainty is sensor fusion~\cite{baltzakis_fusion_2003,liao_parse_2017}, but 
such approaches typically require visual sensors that suffer from limited field 
of view and higher processing time.

We recently shifted the focus from estimating obstacle occupancy from raw 2D 
laser data towards learning to infer \truedist{}s using neural 
networks~\cite{lundell_hallucinating_2018}.
The results demonstrated first of all that typical indoor environments include 
enough 
structure to learn the \truedist{} of objects such as tables and windows, and 
secondly that the learned \truedist{} can improve local navigation safety. 
However, that 
work did not address obstacle position uncertainty, but rather focused on 
situations where raw laser readings could not capture the actual \truedist{} 
and in those cases tried to produce better estimates. In this paper, on the 
other hand, we address the problem of predicting the uncertainty over the deep model used to produce the \truedist{} and show how to integrate this information when building maps.

\paragraph{Mapping and global navigation}
Indoor environments are often represented as occupancy grid maps, where each cell 
represents a portion of the environment that is either free, occupied or 
unknown. When a robot has to plan a path to a particular pose, it searches for 
a trajectory through the free space that reaches the goal position at a minimum 
cost, according to a predefined cost function. Usually, the grid maps 
are built through accumulation of evidence by running the robot in the 
environment. 
This process naturally lends itself to account for uncertainty, as effectively 
each cell contains the belief of it being 
occupied. However, occupancy maps are normally thresholded and used in the 
aforementioned ternary form~\cite{thrun2005probabilistic}. 
Uncertainty is then accounted for in different ways, \eg{} through planning in 
a belief state over obstacle positions to minimize path 
uncertainty~\cite{valencia_path_2011} or by inflating or deflating obstacles 
depending on collision probability~\cite{bicchi_monte_2018}.

 A recent paper 
presented a method for fusing measurement uncertainty directly into the 
map~\cite{vespa_efficient_2018}. That work, however, assumed uncertainty 
stemmed from noisy measurement and not from the inability to detect the 
occupancy of objects in the environment. Whereas in this work it is defined as the uncertainty of a deep model estimating \truedist{}s. This 
enables building uncertainty maps that not only contain actual object 
occupancies but also their positional uncertainty. Here, we present a method for building such maps which is a step forward from ternary occupancy maps.

\subsection{Deep model uncertainty}
\label{sec:uncert}

Standard \ac{dl} architectures provide point estimates, but do not 
inherently capture model uncertainty. Uncertainty can, however, be evaluated 
with \acp{bnn}~\cite{denker1991transforming,mackay_practical_1992} where 
deterministic weights are replaced with distributions over 
the parameters.
Although \acp{bnn} are relatively easy to formulate, using them to perform 
inference is unfeasible as it is often intractable to analytically evaluate the 
marginal probability required for training. Recent new variational inference 
methods have been proposed to address this issue, but they still come with 
increased computational cost, requiring in most cases to double the number of 
parameters of a network to represent its 
uncertainty~\cite{gal_dropout_2016,blundell_weight_2015}.

Another possibility to model uncertainty in \acp{dnn} that has attracted a lot 
of interest recently is to use dropout as approximate Bayesian variational 
inference~\cite{gal_dropout_2016}. The key idea is to enable dropout not only 
in training but also during testing, do several forward passes through the 
network with the same input data, and as a final step estimate the first two 
moments (mean and variance) of the predictive distribution. The mean is then 
used as the estimate, and the variance as a measure of uncertainty. Using this 
approach, researchers have increased semantic segmentation performance on 
images~\cite{kendall_bayesian_2015} and visual relocalization 
accuracy~\cite{kendall_modelling_2016}. Despite the success of using \drop{} to estimate uncertainty in deep models, there are 
scenarios where the approach does not generate reasonable 
results~\cite{osband_deep_2016,osband_risk_2016,pearce_bayesian_2018}. 
In particular, analysis by Osband~\cite{osband_risk_2016} indicates that the 
variance from \drop{} is correlated with the predicted mean. In 
\secref{sec:exp}, we provide 
experimental evidence for this effect, which prevents the use of \drop{} in the 
context of this work, where the impact of the mean on the variance estimate is 
strong.

\section{Method}
\label{sec:method}

\begin{figure*}
	\centering
	\includegraphics[width=.7\linewidth]{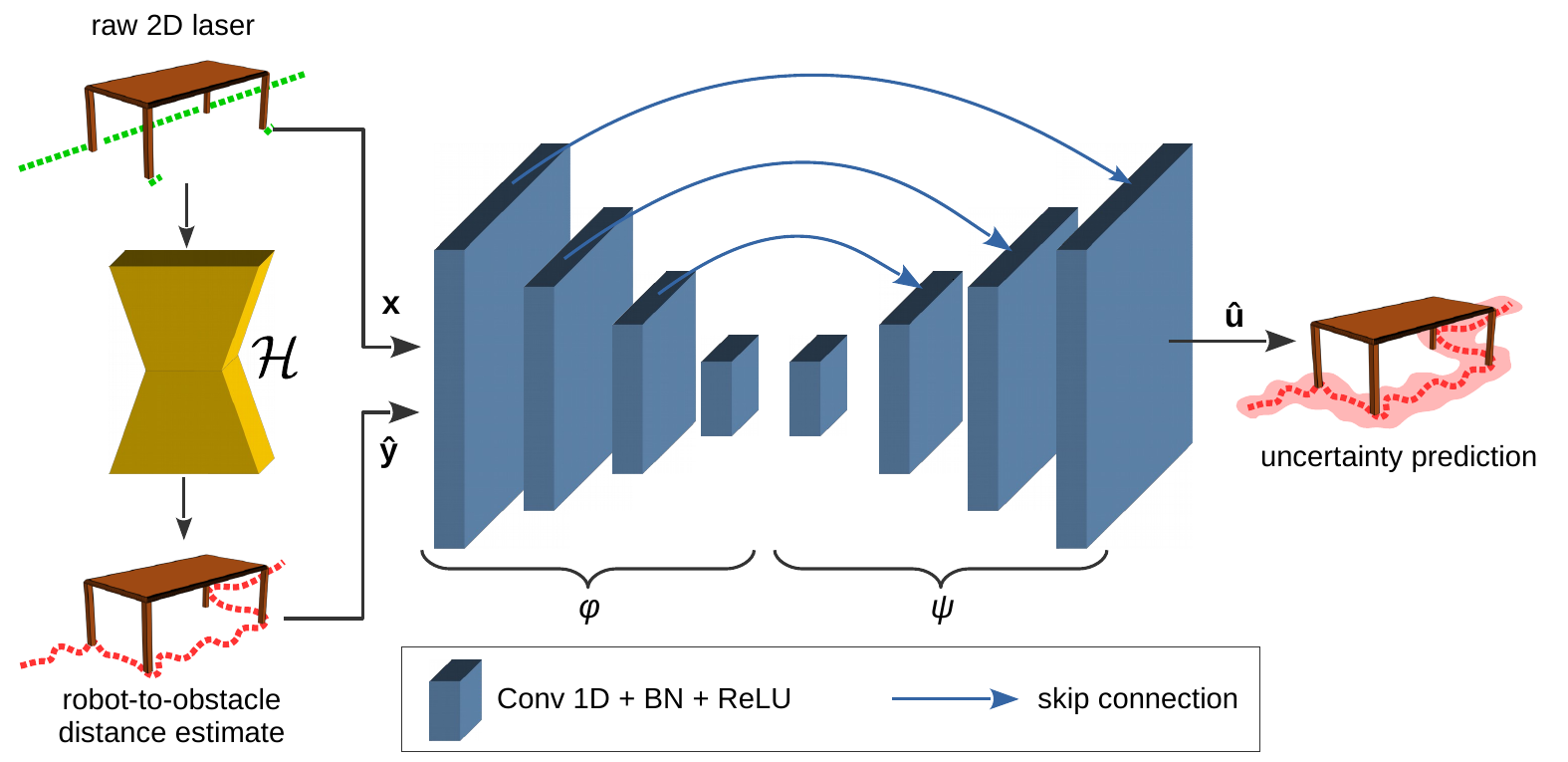}
	\caption{\label{fig:nn}The proposed fully convolutional 
	network~$\mathcal{U}$}
\end{figure*}

In this section, we describe an approach for predicting uncertainty of 
distance estimates and how to integrate it when building 
occupancy maps. Following the formalism presented 
in~\cite{lundell_hallucinating_2018}, let us 
define the output of a generic $N$-point 2D laser positioned at a height~$h$ 
from floor level as a 1D vector $\matr{l}_h = \{l_{ih}\}_{i=1}^{N}$ where 
each~$l_{ih}$ represents an estimate (usually, in meters) of the 
distance~$d_{ih}$ of closest obstacle from the laser at height~$h$, along the 
direction~$i$.
When considering a specific robot with height~$H$ and a 2D laser sensor 
positioned at a fixed height $h^* \in [0,H]$, we define the vector $\matr{x} = 
\matr{l}_{h^*}$.
The \truedist{}s are represented as the closest point along each direction,
\begin{equation} \label{eq:yi}
\matr{y} = \{y_i \mid y_i = \min_{h \in [0,H]} d_{ih}\}_{i=1}^{N}\enspace.
\end{equation}

Let~$\matr{\hat{y}}$ be an estimate of~$\matr{y}$, given from a function 
approximator~$\mathcal{H}$ that takes raw 2D laser signal~$\matr{x}$ as 
input. Such a function approximator was introduced in our original 
work~\cite{lundell_hallucinating_2018}.
In this work, we develop a method to predict the uncertainty 
of the estimator~$\mathcal{H}$. 

\subsection{Uncertainty models}
\label{sec:model}

Assuming independence between each data point, the likelihood that 
$\matr{\hat{y}} = \mathcal{H}(\matr{x})$ follows a parametric model with 
parameters $\theta$ is
\begin{equation}\label{eq:like}
\mathcal{L}(\matr{\hat{y}} \mid \theta) = \prod_{i=1}^N p(\hat{y}_i \mid 
\theta)\enspace.
\end{equation}

We are interested in a model that maximizes~$\mathcal{L}$, or, more 
practically, the log-likelihood $\ell(\matr{\hat{y}} \mid \theta) = \ln 
\mathcal{L}(\matr{\hat{y}} \mid \theta)$.
A common choice is to assume the uncertainty follows a Gaussian model, 
as in the case of \drop{}~\cite{gal_dropout_2016,kendall_what_2017}. 

Under that assumption, one can center the distribution on the 
true \truedist{}~$y_i$ and represent the uncertainty using the model standard 
deviation by finding an uncertainty vector $\matr{\hat{u}} = 
\{\hat{u}_{i}\}_{i=1}^{N}$ such that $\matr{\hat{u}} = \argmax_{\matr{\hat{u}} 
\in \mathbb{R}_{+}^N} \ell_{\mathcal{N}}(\matr{\hat{y}} \mid \matr{y}, 
\matr{\hat{u}})$, where
\begin{equation}\label{eq:loggauss}
\begin{split}
\ell_{\mathcal{N}}(\matr{\hat{y}} \mid \matr{y}, \matr{\hat{u}}) & = 
\sum_{i=1}^N \ln p(\hat{y}_i \mid \mu = y_i, \sigma = \hat{u}_i)\\
& = \sum_{i=1}^N \ln \left( \frac{1}{\sqrt{2 \pi \hat{u}_i^2}} 
\exp\left( - \frac{(\hat{y}_i - y_i)^2}{2 \hat{u}_i^2} \right)\right)\\
& = - \frac{1}{2} \sum_{i=1}^N \left( \ln 2 \pi \hat{u}_i^2 + 
\frac{(\hat{y}_i - y_i)^2}{\hat{u}_i^2} \right)\enspace.
\end{split}
\end{equation}
However, as we will demonstrate in \secref{sec:exp} such a model is unable to
correctly represent the data considered in this work.

Another option is to assume the uncertainty follows a Laplace distribution, a 
choice that was shown to work well for estimating uncertainty in regression 
tasks in vision~\cite{kendall_what_2017}. 
In this case the uncertainty is represented by the scale parameter~$b$ of the 
Laplace distribution by finding $\matr{\hat{u}} = \argmax_{\matr{\hat{u}} \in 
\mathbb{R}_{+}^N} \ell_{L}(\matr{\hat{y}} \mid \matr{y}, \matr{\hat{u}})$, where
\begin{equation}\label{eq:loglaplace}
\begin{split}
\ell_{L}(\matr{\hat{y}} \mid \matr{y}, \matr{\hat{u}}) & = \sum_{i=1}^N \ln 
p(\hat{y}_i \mid \mu = y_i, b = \hat{u}_i)\\
& = \sum_{i=1}^N \ln \left( \frac{1}{2 \hat{u}_i} \exp\left( - 
\frac{|\hat{y}_i - y_i|}{\hat{u}_i} \right)\right)\\
& = - \sum_{i=1}^N \left( \ln 2 \hat{u}_i + 
\frac{|\hat{y}_i - y_i|}{\hat{u}_i} \right)\enspace.
\end{split}
\end{equation}

\subsection{Network architecture}
\label{sec:net}

We propose to train a \ac{dnn}~$\mathcal{U}$ to estimate~$\matr{\hat{u}}$. 
\figref{fig:nn} shows schematically the proposed architecture. The network is 
divided in two components. The first is an encoder~$\varphi$ that takes as 
input a 
2D matrix of length $N = 128$ obtained by concatenating~$\matr{x}$ 
and~$\matr{\hat{y}}$ over the second dimension. It then passes this $128 \times 
2$ input through four 1D convolutional layers (kernel size $K = 5$, stride $S = 
2$), connected via \ac{bn}~\cite{ioffe_batch_2015} and 
\ac{relu}~\cite{he_delving_2015} layers. 
The output sizes of the convolutional layers are, in order, $64\times16$, 
$32\times32$, $16\times64$, and $8\times128$. The output of~$\varphi$ is the 
input to the second part of the network: the decoder~$\psi$, which has the same 
structure as~$\varphi$, and outputs a 1D vector~$\matr{\hat{u}}$ of size~$N$ 
through a series of 1D transposed deconvolutional layers ($K=5$, $S=2$), also 
connected via \ac{bn} and \ac{relu}. The size of the layers' outputs are the 
opposite to the encoder's, \ie{} $16\times64$, $32\times32$, $64\times16$, 
and $128\times1$.
Corresponding convolutional and deconvolutional layers are connected through 
skip connections, as their ability to improve the performance of the 
network has been demonstrated in this domain~\cite{lundell_hallucinating_2018}.
The input of the network is scaled to the range $[0,1]$ to improve convergence 
speed, and weights are optimized using \ac{adam}.
Depending on which uncertainty (Gaussian or Laplacian) the network should 
model, the loss function it minimizes is either the negative log-likelihood of 
\eqref{eq:loggauss} or \eqref{eq:loglaplace}. Although in this work we limit the study to those two models, the proposed 
method can work with any other parametric distribution.

\subsection{Building an uncertainty map}
\label{sec:map}

Next we present a method for building occupancy maps that 
incorporates both the \truedist{} prediction and its uncertainty. To create 
these maps---which we call \textit{uncertainty maps}---we adapt standard 
occupancy mapping techniques~\cite{thrun2005probabilistic} to account for our 
uncertainty estimates similarly 
to~\cite{loop_closed-form_2016,vespa_efficient_2018}. In the following, we will 
present the construction for an uncertainty model following a Laplacian 
distribution; however, the construction for a Gaussian model is similar and is 
discussed in detail in~\cite{vespa_efficient_2018}. 

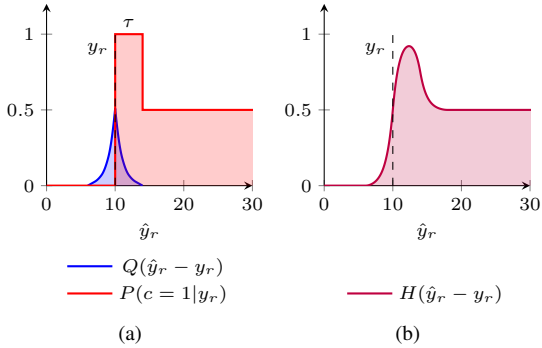
\begin{figure}
	\centering 
	\subfloat[\label{fig:convolution}]{
	\begin{tikzpicture}
		\scriptsize
		\begin{axis}[
			width=.5\linewidth, 
			height=4cm,
			axis x line=bottom,
			axis y line=left,
			xtick distance=10,ytick distance=.5,
			xmin=0, xmax=30,
			ymin=0, ymax=1.2,
			xlabel=$\hat{y}_r$,
			legend style={at={(.5,-.7)},anchor=south,draw=none},
		]
		\addplot[thick, blue, fill=blue, fill opacity=0.2] 
			table {fig/methods/Lspline.txt};
		\addplot[thick, red, fill=red, fill opacity=0.2]
			(0,0)-|(10,1)--(14,1)|-(31,.5)--(31,0);
		\draw[dashed] (10,0)--(10,1);
		\node[anchor=east] at (10,.9) {$y_r$};
		\node[anchor=south] at (12,1) {$\tau$};
		\legend{$Q(\hat{y}_r-y_r)$, {$P(c = 1|y_r)$}};
		\end{axis}
	\end{tikzpicture}}
	\subfloat[\label{fig:res_convolution}]{
	\begin{tikzpicture}
		\scriptsize
		\begin{axis}[
			width=.5\linewidth, 
			height=4cm,
			axis x line=bottom,
			axis y line=left,
			xtick distance=10,ytick distance=.5,
			xmin=0, xmax=30,
			ymin=0, ymax=1.2,
			xlabel=$\hat{y}_r$,
			legend style={at={(.5,-.7)},anchor=south,draw=none},
		]
		\addplot[thick, purple, fill=purple, fill opacity=0.2] 
		table {fig/methods/occup.txt}--(31,.5)--(31,0);
		\draw[dashed] (10,0)--(10,1);
		\node[anchor=east] at (10,.9) {$y_r$};
		\legend{$H(\hat{y}_r-y_r)$};
		\end{axis}
	\end{tikzpicture}}
	\caption{\label{fig:grid_example}The geometric construction of 
	$P(c = 1|y_r)$, where~$\tau$ equals half the support of~$Q$ (a), and the 
	resulting convolution (b).}
\end{figure}

The occupancy probability $P(c=1|\hat{y}_r)$ of 
a cell~$c$ given an estimated \truedist{}~$\hat{y}_r$ along a laser ray~$r$ 
is defined as
\begin{equation}
\label{eq:occProbInt}
P(c=1|\hat{y}_r)=\int_{0}^{\infty}P(c=1|y_r)P(y_r|\hat{y}_r)\de{y_r}\enspace,
\end{equation}
where~$y_r$ represents the true \truedist{} along ray~$r$. Here, we 
model the occupancy given the true \truedist{} $P(c=1|y_r)$ as $0$ in 
front of the obstacle, $1$ from the surface to~$\tau$ units behind, and then 
$0.5$. An example of $P(c=1|y_r)$ is shown in red in \figref{fig:convolution}.

\begin{figure}
	\centering
	\begin{tikzpicture}
	\scriptsize
	\begin{axis}[
	width=\linewidth, 
	height=4cm,
	axis x line=bottom,
	axis y line=left,
	xtick distance=2,ytick distance=.1,
	xmin=0, xmax=17,
	ymin=0, ymax=0.59,
	xlabel=$\hat{y}$,
	ylabel=,
	legend pos=north west,
	legend style={draw=none},
	legend cell align={left},
	]
	\addplot[thick, blue] table {fig/methods/Lspline.txt};
	\addplot[domain=0:10, thick, dashed, red, samples=100] 
	{.5 * exp(-abs(x-10))};
	\addplot[domain=10:18, thick, dashed, red, samples=100, forget plot]
	{.5 * exp(-abs(x-10))};
	\legend{$Q(\hat{y}-10)$, {$\operatorname{Laplace}(\mu = 10, b = 1)$}};
	\end{axis}
	\end{tikzpicture}
	\caption{\label{fig:spline}The quadratic B-spline function~$Q$, 
		and the corresponding Laplace}
\end{figure}
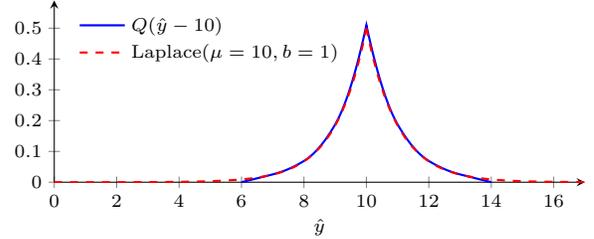

Calculating the integral in \eqref{eq:occProbInt} requires a model 
$P(y_r|\hat{y}_r)$ of the true \truedist{} given a noisy measurement reading. 
As stated in \secref{sec:method}, $P(y_r|\hat{y}_r)$ 
is modeled as a Laplace distribution. However, as pointed out 
in~\cite{loop_closed-form_2016}, integrating \eqref{eq:occProbInt} over a 
distribution with infinite support results in obstacles lying slightly behind the range 
measurement even when the measurements are precise. One solution is to 
set $\tau = \infty$, but this can degenerate occupancy 
probabilities~\cite{loop_closed-form_2016}. Another, more stable option, is to 
limit the support of $P(y_r|\hat{y}_r)$. In this work, similarly 
to~\cite{loop_closed-form_2016}, the support of a 
Laplace distribution is limited by approximating it as a quadratic B-spline 
$Q(t) \approx P(y_r|\hat{y}_r)$ where $t = (y_r-\hat{y}_r) / \hat{u}_r$ 
and 
$\hat{u}_r = \mathcal{U}(x_r, \hat{y}_r)$. We derived $Q$ numerically by 
imposing it to have finite support $[-4, 4]$ and unit integral. The choice of 
the range for the support originates from the fact that, for a 
$\operatorname{Laplace}(0,1)$, around 99\% of the cumulative distribution lies 
in that range. \figref{fig:spline} shows the resulting spline approximation.

By choosing the thickness term $\tau = 4\hat{u}_r$, \ie{} half the support of 
the approximated B-spline, \eqref{eq:occProbInt} has the analytical solution
\begin{equation}\label{eq:occProbMeas}
P(c=1|\hat{y}_r) \approx 
H(t)=Q_\text{cdf}(t)-\frac{1}{2}Q_{\text{cdf}}(t-4)\enspace,
\end{equation}
where~$Q_\text{cdf}(t)$ is the cumulative density function of the quadratic 
B-spline~$Q(t)$. The per-ray occupancy probability is visualized in 
\figref{fig:res_convolution}. 

\begin{algorithm}[t]
	\caption{\label{alg:uncertainty_map}Uncertainty Map}
	\begin{algorithmic}
		\STATE \textbf{Inputs:} $\matr{P}$: Robot poses, $\matr{L}$: 
		Laser scans,\\
		\bindent
		\STATE $\matr{M}$: Empty map, $\alpha$: Correlation factor. 
		\eindent
		\STATE			
		\FORALL{$p \in \matr{P}$}
		\STATE $\matr{x} \leftarrow \textsc{GetLaserScan}(\matr{L}, p)$
		\STATE $\matr{\hat{y}} \leftarrow \mathcal{H}(\matr{x})$
		\STATE $\matr{\hat{u}} \leftarrow \mathcal{U}(\matr{x},
		\matr{\hat{y}})$
		\FORALL{$r \in 1, \dots, N$}
		\STATE $\matr{C} \leftarrow \textsc{RayCast}(p, \hat{y}_r)$
		\FORALL{$(c \in \matr{C})$}
		\STATE $d_{c\hat{y_r}} \leftarrow \textsc{EuclidDist}(c,\hat{y}_r)$  \\
		\STATE $o_c \leftarrow H(\frac{d_{c\hat{y}_r}-\hat{y}_r}{\hat{u}_r})$\\
		\STATE $m_c \leftarrow \log\frac{o_c}{1-o_c}$\\
		\STATE $\matr{M}(c) \leftarrow \matr{M}(c) + \alpha m_c$
		\ENDFOR
		\ENDFOR
		\ENDFOR
		\RETURN $\frac{1}{1+\exp(\matr{M})}$
	\end{algorithmic}
\end{algorithm}

The method for creating uncertainty maps with occupancy probabilities calculated in 
\eqref{eq:occProbMeas} is summarized in \algoref{alg:uncertainty_map}. We 
assume robot poses~$\matr{P}$ and original laser scans~$\matr{L}$ to be given. 
First for each robot pose~$p$, the algorithm extracts the associated original 
laser scan~$\matr{x}$. This laser scan is then propagated through 
network~$\mathcal{H}$ to estimate the \truedist{}~$\matr{\hat{y}}$ which, 
together with the original laser, is propagated through~$\mathcal{U}$ to 
estimate the scale parameters~$\matr{\hat{u}}$ of the Laplacian uncertainty 
model. Next, the algorithm goes through every laser ray~$\hat{y}_r$ in the 
current scan~$\matr{\hat{y}}$ and finds all cells~$\matr{C}$ that are touched 
by the laser ray~$r$ originating from the current pose~$p$. 
Then the algorithm iterates over each cell $c \in \matr{C}$, calculates the 
distance~$d_{c\hat{y}_r}$ between the center of the cell and the estimated 
obstacle distance~$\hat{y}_r$, calculates the occupancy for that cell using 
\eqref{eq:occProbMeas} with $t=(d_{c\hat{y}_r}-\hat{y}_r) / \hat{u}_r$, 
converts the occupancy to \textit{log-odds}, scales it with a 
correlation factor~$\alpha$, and finally adds it to the previous occupancy 
value for that cell. As a final step the uncertainty map~$\matr{M}$ is 
converted from log-odds to occupancy probability, \ie{} in $[0,1]$. 

The correlation factor $0<\alpha<1$ limits the effect for which highly 
correlated readings could produce overconfident maps and remove most of the 
uncertainty. This parameter should be set as a function of the amount of 
correlation between successive readings; in this work we experimentally 
set it to a constant value $\alpha=0.01$ as we are working in a static 
environment. In a dynamic environment, $\alpha$ can be set as a function of the 
time passed between readings to have the map adapt to changes in the 
environment~\cite{vespa_efficient_2018}.
\section{Experiments}
\label{sec:exp}

\begin{table}
	\centering
	\caption{\label{tab:loglike}Average Log-likelihood over the validation sets}
	\begin{tabular}{llrr}
		\toprule
		Algorithm & Model & \valI{} & \valII{}\\ 
		\midrule
		MC Dropout & Gaussian & -2.55 & -4.63\\ 
		Ours & Gaussian & -1.42 & 1.48\\ 
		Ours & Laplace & \textbf{1.65} & \textbf{3.25}\\
		\bottomrule 
	\end{tabular} 
\end{table}

\begin{figure}
	\centering
	\subfloat[\label{fig:plotsYUh}Ours 
	(Laplace)]{\includegraphics[width=.45\linewidth]{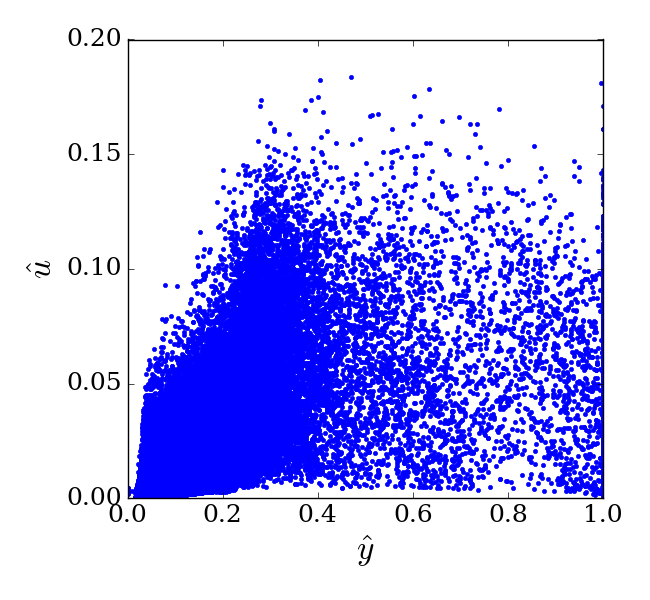}}
	\subfloat[\label{fig:plotsYUhd}\drop{}]{\includegraphics[width=.45\linewidth]{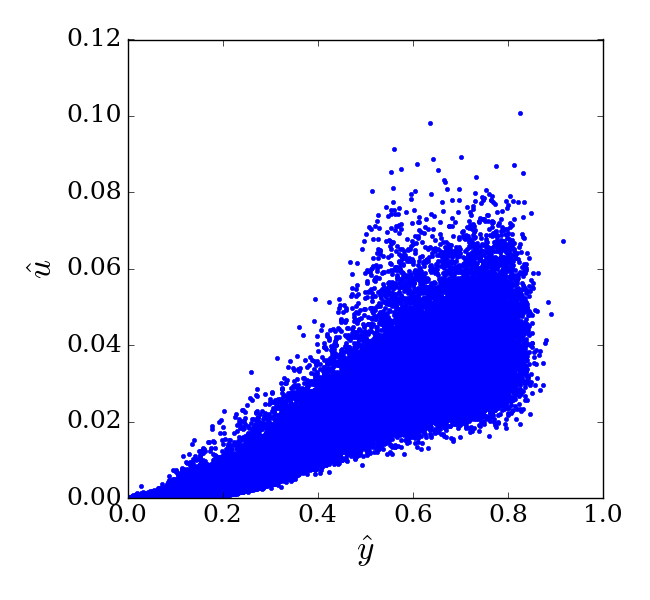}}
	\caption{\label{fig:plots}Plots of the relationship between the estimated 
		\truedist{}~$\matr{\hat{y}}$ and uncertainty~$\matr{\hat{u}}$ for the 
		proposed method and \drop{}.}
\end{figure}

\begin{figure*}
	\centering
	\subfloat[\label{fig:slam_gt}SLAM]
	{\includegraphics[width=.242\linewidth]{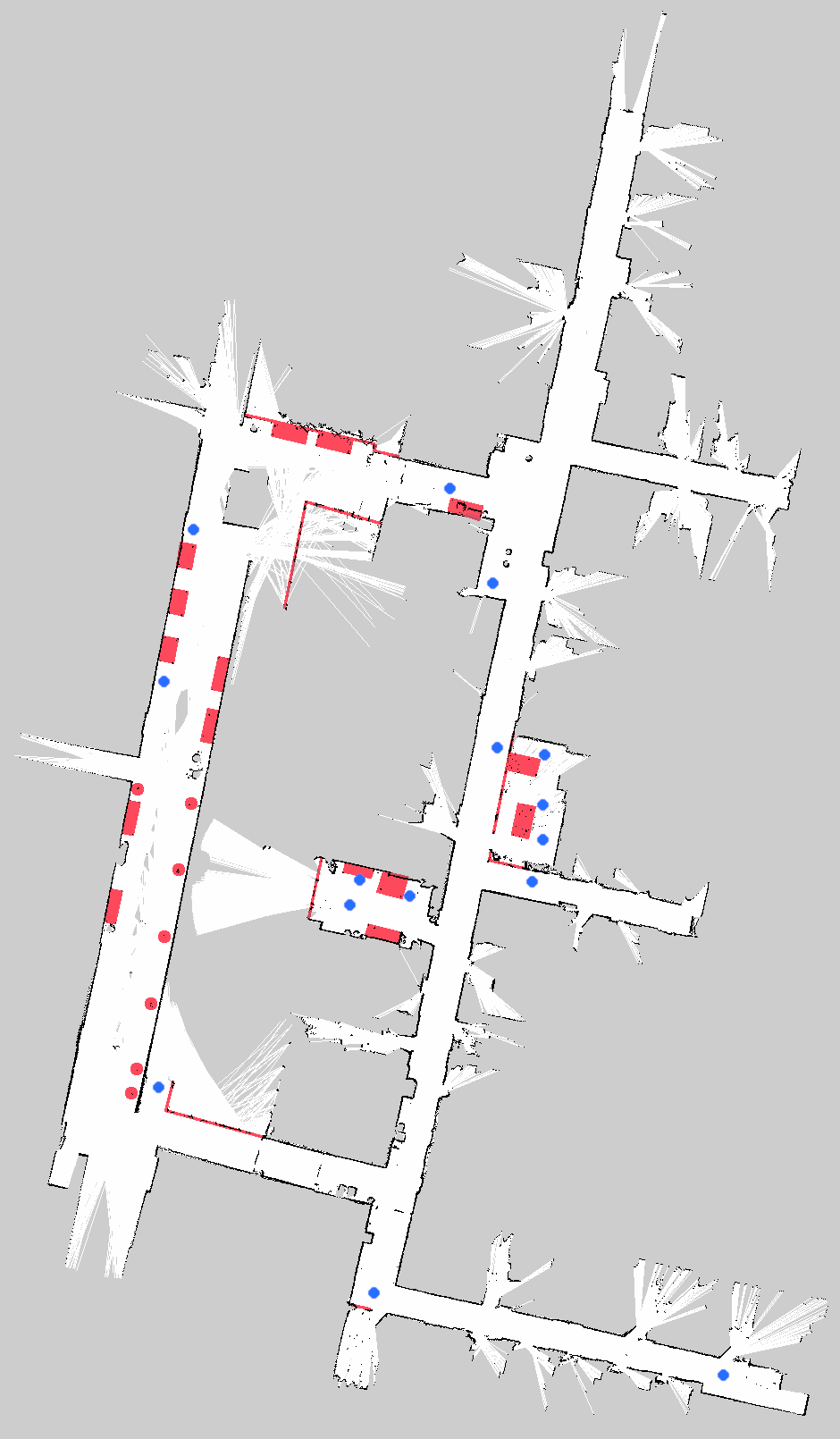}}~
	\subfloat[\label{fig:hallucinated_uncertainty_dropout}\drop{}]
	{\begin{overpic}[width=.242\linewidth]{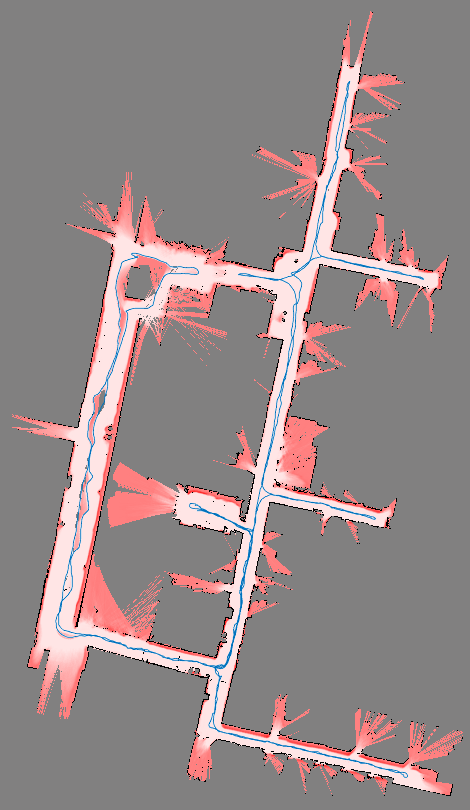}
		\linethickness{0.5mm}\color[HTML]{3ddc97}%
		\put(16,43){\scriptsize\figref{fig:office}}%
		\put(20,38){\polygon(0,3)(11,3)(11,-5)(0,-5)}%
		%
	\end{overpic}}~
	\subfloat[\label{fig:hallucinated_uncertainty_gauss}Ours (Gaussian)]
	{\includegraphics[width=.242\linewidth]{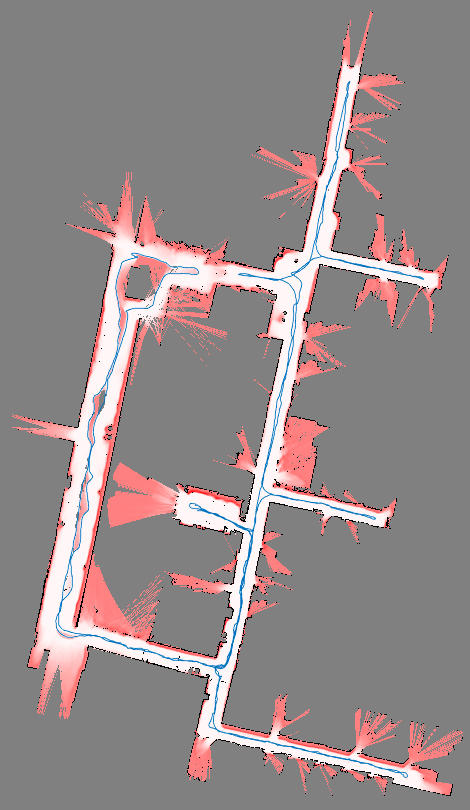}}~
	\subfloat[\label{fig:hallucinated_uncertainty_laplace}Ours (Laplace)]
	{\includegraphics[width=.242\linewidth]{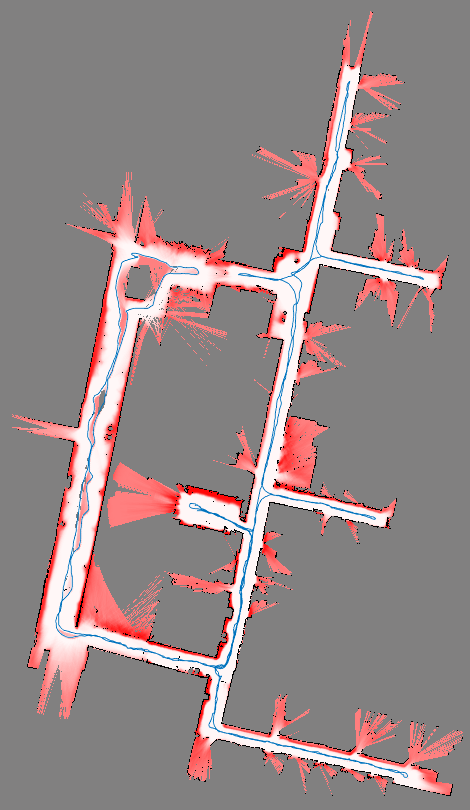}}
	\caption{\label{fig:maps}The maps created by gathering data with a \cob{}.
		In (a) the obstacles invisible to the laser scanner are indicated in 
		red, and the 15 navigation goals used in the 
		navigation experiment in blue. In all other figures, the original 
		\ac{slam} map is visualized in gray scale while the superimposed 
		uncertainty maps shown in degrees of red. As in regular occupancy maps, 
		darker red indicates a higher occupancy probability. The blue line is 
		the robot path \bestcolor{}.}
\end{figure*}

In this work, we propose a framework for building uncertainty maps based on 
deep network uncertainty models. Evaluation of the proposed approach is then 
two-fold: first of all we are interested in assessing the quality of the 
proposed uncertainty model (\secref{sec:exp_model}); secondly, we want to 
analyze the resulting uncertainty maps to validate the suggested mapping 
strategy (\secref{sec:exp_map}). However, for completeness, we first present the 
network training setup.

\subsection{Dataset and network training}
\label{sec:data}

We used the trained network~$\mathcal{H}$, whose training setup was detailed 
in~\cite{lundell_hallucinating_2018}.
Then, to train the uncertainty network~$\mathcal{U}$, we took the complete 
training and test set used in~\cite{lundell_hallucinating_2018}, available 
online\footnote{\href{https://github.com/jsll/IROS2018-Hallucinating-Robots}
{github.com/jsll/IROS2018-Hallucinating-Robots}}, and merged them. These two 
datasets were gathered in a university building, on the second and third floor, 
respectively, and once merged consisted of 39508 pairs of laser scan~$\matr{x}$ 
(input of~$\mathcal{H}$) and relative \truedist~$\matr{y}$ (ground-truth 
output).
We propagated each laser scan~$\matr{x}$ through the trained 
network~$\mathcal{H}$, saved each pair $(\matr{x}, \matr{\hat{y}})$ as training 
input for~$\mathcal{U}$, and kept~$\matr{y}$ as ground-truth data to 
compute the loss as explained in \secref{sec:net}. We saved these new 
input-output pairs as a new dataset we used to train network~$\mathcal{U}$. The 
reason for merging the two datasets is that the errors produced 
by~$\mathcal{H}$ over the training set differs slightly compared to the test 
set, due to the training process; thus, merging the two datasets enable us to 
capture a wider range of uncertainties while training~$\mathcal{U}$ while 
limiting overfitting. The network was trained for 2000 epochs, with a batch 
size of 32, and learning rate set to~$10^{-4}$.

For validation, we gathered new data by teleoperating a \cob{} in the same 
environments where the previous datasets were gathered. This resulted in two 
new datasets, which we will refer to as \valI{} and \valII{}. \valI{} consists 
of 18961 samples gathered on the second floor while \valII{} consists of 24829 
samples from the third floor. By performing a new acquisition we ensure that we 
conduct our evaluation on data which is uncorrelated with the training set.

\subsection{Model quality evaluation}
\label{sec:exp_model}

To evaluate the quality of the proposed uncertainty model we compare its 
performance to that obtained by \sota{} \drop{}~\cite{gal_dropout_2016}.
To obtain \drop{}'s results we modified the network 
in~\cite{lundell_hallucinating_2018} to include dropout layers after each 
convolutional layer. We will refer to this modified version of~$\mathcal{H}$ 
as~$\mathcal{H}_D$. We then trained the network with a dropout rate of 50\%. 
In testing, to evaluate uncertainty, we performed 50 forward passes of the same 
input in the network using the same dropout rate.
We then took the average of this sampling process as the network prediction for 
the \truedist{}~$\matr{y}$ and the variance of the samples as the estimated 
uncertainty~$\matr{\hat{u}}$, as proposed in~\cite{gal_dropout_2016}.

\tabref{tab:loglike} shows average log-likelihood scores over the two 
validation sets for our approach, both using Gaussian and Laplacian models, 
and \drop{}. It is noticeable that both methods using Gaussian model 
are unable to correctly model uncertainty of \truedist{}s and get considerably 
lower scores than our approach using Laplacian model. This makes it
impossible to use an approach like \drop{} for the application at hand; given 
that it assumes by design the uncertainty to be Gaussian. Our approach instead, 
given its ability to be optimized for any distribution, can be used 
effectively, \eg{} by selecting a Laplacian model.

Moreover, \figref{fig:plots} presents plots examining the relationship between 
the estimated \truedist{}~$\matr{\hat{y}}$ (horizontal axis) and the relative 
predicted uncertainty~$\matr{\hat{u}}$ (vertical axis) for both our approach 
and \drop{}. 
Each plot represents results over the whole \valI{} set where each point 
corresponds to one lidar distance reading.
It is apparent from \figref{fig:plotsYUhd} that for \drop{} a strong 
correlation between~$\matr{\hat{u}}$ and~$\matr{\hat{y}}$ is manifested. Given 
the lidar sensor, this should not be the case: the uncertainty should be 
independent of the distance at which the obstacle lays. As seen in \figref{fig:plotsYUh}, no similar effect is 
present for our model. This side-effect of \drop{}, 
which was theoretically noted in~\cite{osband_risk_2016}, is therefore 
confirmed by our empirical evaluation, demonstrating that the use of \drop{} 
may not be suitable for domains, like the one at hand, where there is a wide 
range of possible output values and where therefore the dependence of \drop{}'s 
variance from the estimated mean overshadows any other contribution.

\begin{figure}
	\subfloat[\label{fig:office_drop}\drop{}]{
		\begin{overpic}[width=.482\linewidth]{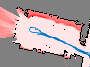}
			\linethickness{0.5mm}\color[HTML]{3ddc97}%
			\put(75,62){Tables}%
			\put(30,60){\polygon(0,10)(60,-5)(60,-28)(0,-13)}%
			\put(2,10){Windows}%
			\put(10,30){\polygon(0,40)(15,40)(15,-10)(0,-10)}%
			
	\end{overpic}}~
	\subfloat[\label{fig:office_our}Ours 
	(Laplace)]{\includegraphics[width=.482\linewidth]{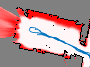}}
	\caption{\label{fig:office}Marked area in 
	\figref{fig:hallucinated_uncertainty_dropout} \bestcolor{}.}
\end{figure}

\subsection{Uncertainty maps}
\label{sec:exp_map}

Building an uncertainty map requires robot poses, laser scans, and the inverse 
measurement model. To collect these we first teleoperated a \cob{} in our  
facility to gather odometry and laser range data, then we built the \ac{slam} 
map shown in \figref{fig:maps} using ROS gmapping, and finally we ran 
ROS amcl\footnote{gmapping: \url{http://wiki.ros.org/gmapping},\\amcl: 
\url{http://wiki.ros.org/amcl}} to register robot poses depicted as the blue 
path in \figref{fig:maps}. 

Once laser measurements and odometry were collected, we created, in addition 
to the base \ac{slam} map, three other maps: a map obtained using \drop{} 
(\figref{fig:hallucinated_uncertainty_dropout}), and two different maps using 
our proposed approach, one modeling the uncertainty as a
Gaussian distribution (\figref{fig:hallucinated_uncertainty_gauss}) and 
another one modeling it as a Laplace
(\figref{fig:hallucinated_uncertainty_laplace}). All three maps were created 
using \algoref{alg:uncertainty_map} with $\alpha=0.01$, where the only 
difference were in determining~$\matr{\hat{u}}$. In the maps depicted in 
\figsref{fig:hallucinated_uncertainty_gauss}{fig:hallucinated_uncertainty_laplace}~$\matr{\hat{u}}$
 was set as the output from the uncertainty network, while for the \drop{} map 
 in \figref{fig:hallucinated_uncertainty_dropout}~$\matr{\hat{u}}$ was set as the standard 
deviation of the outputs obtained from 50 forwards passes of the same input 
through network~$\mathcal{H}_D$.

A magnified sections of these maps are shown in 
\figref{fig:office}. From visually inspecting the figure one 
can conclude that the uncertainty maps (visualized as the red shaded area) are 
coherent with the original \ac{slam} map in areas with low uncertainty such as 
for walls, while differing in less certain areas which includes windows and tables.
Although our Gaussian uncertainty map and the \drop{} map are very 
similar, the main difference is that the \drop{} map is more uncertain 
about cells that are clearly free including the robot path. 

With regards to the map in \figref{fig:hallucinated_uncertainty_dropout} 
produced by \drop{}, it is worth noticing that although the map looks 
reasonable thanks to the good estimate of the distance, the estimate of the 
uncertainty in that map is inaccurate, as demonstrated in 
\secref{sec:exp_model}. 
This causes the map to underestimate the occupancy of obstacles: given 
the correlation between the uncertainty and the estimated distance (see 
\figref{fig:plotsYUhd}): when the robot is 
very close, any estimate is taken as certain. This is apparent in 
particular in \figref{fig:office}, where the \drop{} map cannot 
capture the occupancy of tables nor windows, which our Laplace 
map estimates correctly.

\subsection{Global navigation on uncertainty maps}

To test the ability of uncertainty maps to improve global navigation safety, we 
run a simulation experiment where we left a robot randomly navigate an 
environment with and without uncertainty maps. More precisely, we setup a 
Gazebo simulation mirroring the real \valI{} environment, we then had 
a robot plan a sequence of 400 trajectories each connecting two points from a
set of 15 goal points distributed over the environment\footnote{The code for 
the navigation experiment is available for download at 
\href{https://github.com/aalto-intelligent-robotics/uncertain_turtlebot_navigation}
{github.com/aalto-intelligent-robotics/uncertain\_turtlebot\_navigation}}. 
The path planning was performed by using standard ROS gmapping. We repeated the 
experiment changing each time the costmap used by gmapping to include the 
\ac{slam} map and either one of the three uncertainty maps in 
\figref{fig:maps}. As baseline, we ran the same navigation using only a \ac{slam} map without uncertainty. We manually marked in the environment occupancy of objects 
invisible to the laser (\eg{} table, benches and glass walls) and then computed 
the number of time a navigation path would collide with any of these obstacles 
when using each of the four different maps. \figref{fig:slam_gt} shows the 15 
goal points in blue and the aforementioned occupied areas in red.

\begin{table}
	\centering
	\caption{\label{tab:nav}Number of path resulting in collisions over 400 
		trajectories for the different types of map}
	\begin{tabular}{llrr}
		\toprule
		Map & Model & n. of collision & Percentage\\ 
		\midrule
		\ac{slam} only & no uncertainty & 23 & 5.75\%\\
		\drop{} & Gaussian & 76 & 19.00\%\\ 
		Ours & Gaussian & 71 & 17.75\%\\ 
		Ours & Laplace & \textbf{0} & \textbf{0.00\%}\\
		\bottomrule 
	\end{tabular} 
\end{table}

\tabref{tab:nav} presents the results of this experiment, which
demonstrates the benefits uncertainty maps can have on global navigation 
safety. However, the model of the uncertainty needs to be accurate, otherwise 
it might mislead the planner. As we discussed in the previous sections, 
Gaussian models produce overconfident predictions which resulted in even 
more collision compared to planning on \ac{slam} maps alone. On the other hand, 
the ability of our Laplace model to correctly estimate areas of high 
uncertainty enabled the robot to avoid risky areas and prefer safer options, 
resulting in safer navigation and, in our experiment, no collisions.

\begin{figure}
	\centering
	\subfloat[\label{fig:nav_slam}SLAM]{\includegraphics[width=.482\linewidth]{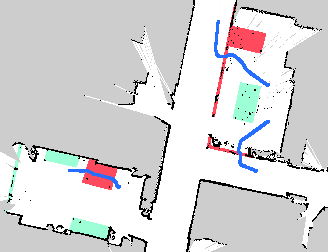}}~
	\subfloat[\label{fig:nav_mcdropout}\drop{}]{\includegraphics[width=.482\linewidth]{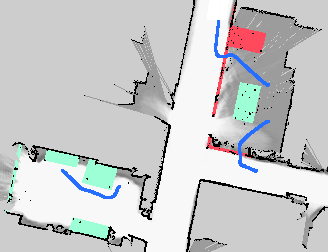}}\\
	\subfloat[\label{fig:nav_gaussian}Ours 
	(Gaussian)]{\includegraphics[width=.482\linewidth]{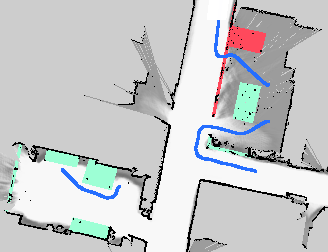}}~
	\subfloat[\label{fig:nav_laplace}Ours 
	(Laplace)]{\includegraphics[width=.482\linewidth]{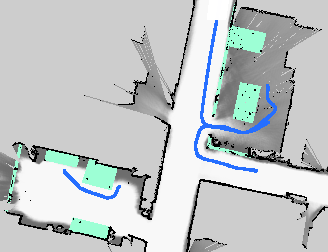}}
	\caption{\label{fig:nav}Three sample navigation trajectories (blue lines) 
		executed on four different maps. The true occupancy of objects 
		invisible to 
		the laser is overlaid on the maps, in green when no collision occurred 
		or 
		in red in case of collision. The uncertainty map is shown in shades of 
		gray 
		\bestcolor{}.}
\end{figure}

\figref{fig:nav} shows an example of these trajectories. It is apparent that 
the path planned in our Laplace map (see \figref{fig:nav_laplace}) ends in the 
correct position without any risk for collisions as it plans around more 
uncertain areas, keeps further away from the walls, and successfully avoids the 
glass surface. The same is not true for all paths planned on the \ac{slam} map 
(\figref{fig:nav_slam}) and at least some of the paths planned on the other 
uncertainty maps (\figsref{fig:nav_mcdropout}{fig:nav_gaussian}).
\section{Conclusions}
\label{sec:concl}

Building occupancy maps in environments full of complex obstacles is 
challenging but crucial for autonomous robot navigation.
To this end, we presented a method for predicting model uncertainty of a deep 
network estimating \truedist{} and proposed a mapping algorithm to create 
uncertainty 
maps for indoor autonomous global navigation. To achieve this we trained an 
uncertainty network that takes both estimated and original laser readings as 
inputs and outputs an uncertainty measure over the estimated distances. In the 
quantitative evaluation when the uncertainty network modeled uncertainties as 
Laplacian opposed to Gaussian it produced much better predictions in comparison 
to the \sota{} \drop{}, indicating that wider tailed distributions are better 
equipped to capture uncertainties in this domain. Extension of this idea to 
other domains is left as future work.

The ability of the Laplacian model to better represent obstacle distances is 
also apparent when creating uncertainty maps, where the map built 
using our Laplacian strategy successfully marked otherwise invisible objects 
such as glass as occupied, while the map built using a Gaussian model did not 
capture such areas due to the overconfident uncertainty estimates. In addition, 
we demonstrated that paths planned on an uncertainty map prefer areas of lower 
uncertainty and reduce the probability of collision, increasing autonomous 
navigation safety over the use of traditional ternary occupancy maps.

Finally it is worth mentioning that although the proposed architecture was only 
tested in this domain, it is agnostic to the estimator~$\mathcal{H}$ and to the 
data it is processing, indicating that it may generalize to model uncertainties 
in other domains as well. However, generalization of this architecture is out 
of the scope of this study and is left as future work.


%



\bibliographystyle{IEEEtran}
\bibliography{refs}

\end{document}